\documentclass[10pt,twocolumn,letterpaper]{article}

\usepackage{cvpr}              
\usepackage{multirow}

\usepackage[dvipsnames]{xcolor}

\definecolor{cvprblue}{rgb}{0.21,0.49,0.74}
\usepackage[pagebackref,breaklinks,colorlinks,citecolor=cvprblue]{hyperref}

\def\paperID{11577} 
\def\confName{CVPR}
\def\confYear{2024}

\title{ManipLLM: Embodied Multimodal Large Language Model for \\ Object-Centric Robotic Manipulation }

\author{
Xiaoqi Li\thanks{clorisli@stu.pku.edu.cn}, 
Mingxu Zhang,
Yiran Geng,
Haoran Geng,
Yuxing Long,\\
Yan Shen,
Renrui Zhang,
Jiaming Liu,
Hao Dong\thanks{Corresponding author: hao.dong@pku.edu.cn}
\vspace{0.1cm}\\
School of Computer Science, Peking University   
}

\begin{document}
\maketitle
\begin{abstract}
Robot manipulation relies on accurately predicting contact points and end-effector directions to ensure successful operation. 
However, learning-based robot manipulation, trained on a limited category within a simulator, often struggles to achieve generalizability, especially when confronted with extensive categories.
Therefore, we introduce an innovative approach for robot manipulation that leverages the robust reasoning capabilities of Multimodal Large Language Models (MLLMs) to enhance the stability and generalization of manipulation. 
By fine-tuning the injected adapters, we preserve the inherent common sense and reasoning ability of the MLLMs while equipping them with the ability for manipulation. 
The fundamental insight lies in the introduced fine-tuning paradigm, encompassing object category understanding, affordance prior reasoning, and object-centric pose prediction to stimulate the reasoning ability of MLLM in manipulation. 
During inference, our approach utilizes an RGB image and text prompt to predict the end effector's pose in chain of thoughts. 
After the initial contact is established, an active impedance adaptation policy is introduced to plan the upcoming waypoints in a closed-loop manner. 
Moreover, in real world, we design a test-time adaptation (TTA) strategy for manipulation to enable the model better adapt to the current real-world scene configuration. 
Experiments in simulator and real-world show the promising performance of ManipLLM.
More details and demonstrations can be
found at \href{https://sites.google.com/view/manipllm}{https://sites.google.com/view/manipllm}.
\end{abstract} 
    
\section{Introduction}
\label{sec:intro}

\begin{figure}[t]
\begin{center}
    \includegraphics[width=\linewidth]{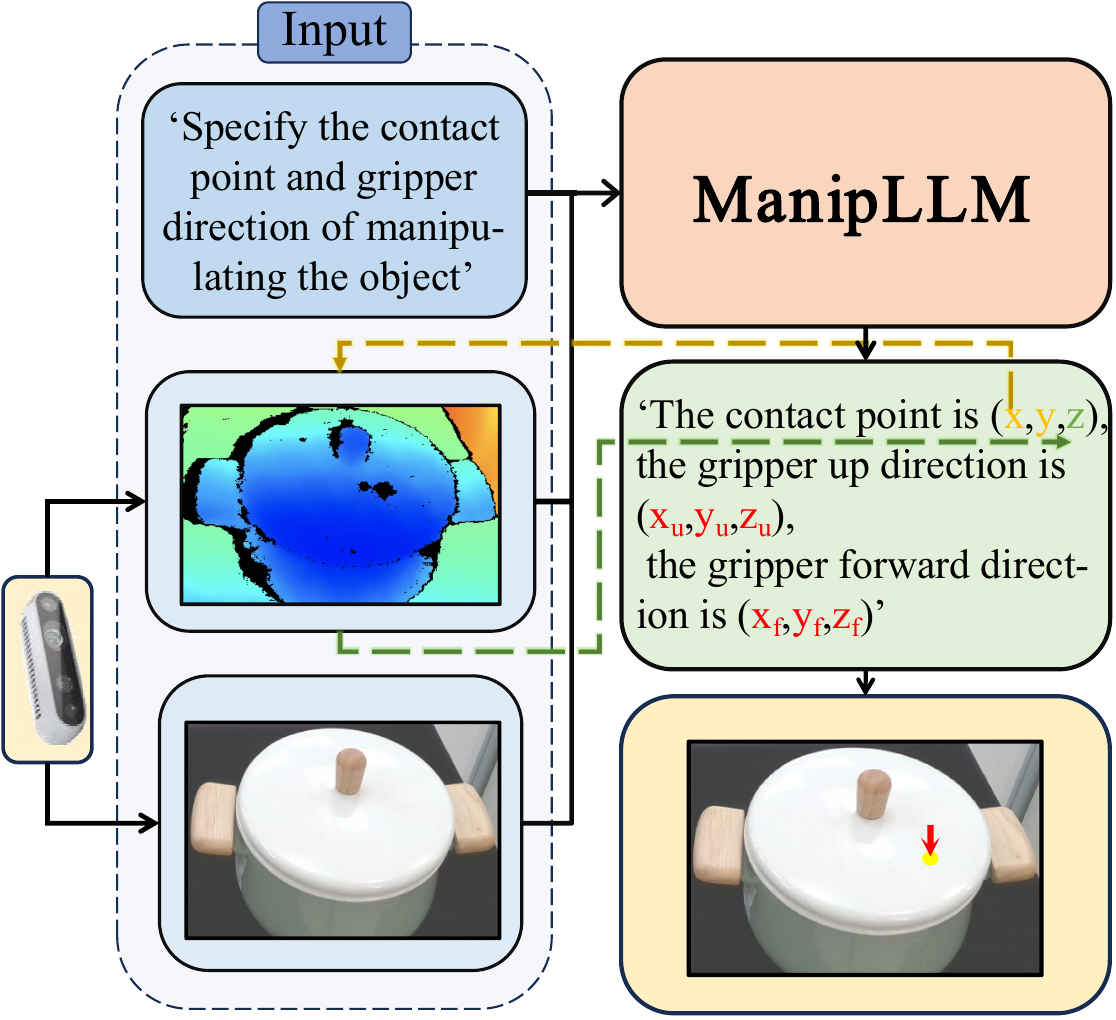}
\end{center}
\caption{The prediction of ManipLLM. Given the text prompt, RGB image, and depth map inputs, we obtain 3D contact point $(x, y, z)$. Here, $x$ and $y$ represent the pixel coordinates in the image predicted by ManipLLM, while $z$ corresponds to the depth obtained from the depth camera. Additionally, ManipLLM predicts the gripper's up direction $(x_u, y_u, z_u)$ and forward direction $(x_f, y_f, z_f)$, forming the end-effector $SO(3)$ rotation.}
\label{fig:teasor}
\end{figure}


As robot manipulation requires robots to interact with diverse objects, the robustness and explainability of low-level action prediction become essential for manipulating reliability.
While certain approaches~\cite{mao2023gpt,Mo_2019_CVPR,geng2022end,geng2023partmanip,geng2023rlafford,geng2023gapartnet} demonstrate impressive performance, they often sacrifice interpretability by treating low-level manipulation prediction as a black-box prediction problem and lack the inherent common-sense reasoning abilities inherent in humans, limiting their capacity to manipulate wide-spread categories of objects.

Existing advancements in Multimodal Large Language Models (MLLMs)\cite{mao2023gpt,li2022blip,alayrac2022flamingo,zhang2023llama} highlight their proficiency in common sense reasoning and remarkable generalization in vision tasks~\cite{anderson2018bottom,fukui2016multimodal}. However, training Multi-Label Language Models (MLLMs) directly to learn robotic low-level action trajectories (\ie end-effector trajectories)~\cite{brohan2022rt,zitkovich2023rt} poses challenges in generalization due to minimal low-level action samples in their pretraining data. Consequently, MLLMs lack prior knowledge in this field while successful training for these tasks necessitates extensive data to achieve desired generalization ability.
In contrast, MLLMs exhibit robust capabilities in comprehending objects and demonstrate significant generalization abilities. Given these considerations, transforming MLLMs to object-centric manipulation proves more efficient. This then raises an important question: How can we harness MLLMs to facilitate object-centric robot manipulation?
The major challenge is how to enable MLLMs to understand the geometric structure of objects (such as their axis) to predict the movable contact positions for object-centric manipulation. Furthermore, it remains unexplored whether these models, which take 2D inputs, can also predict 3D end-effector directions.


In this study, we aim to exploit the common sense and reasoning ability embedded within MLLMs~\cite{zhang2023llama,touvron2023llama} to realize promising robot manipulation performance.
%
To accomplish this, during training, in order to preserve the powerful ability of MLLMs and empower them with manipulation ability, we only finetune the injected learnable adapters~\cite{hu2021lora} on MLLMs.
Furthermore, we design an intricate training paradigm and formulate fine-tuning tasks, including object category identification, affordance prior reasoning, and manipulation-aware pose prediction. The affordance prior considers the geometric intrinsics of the object and reflects the probability of generating movement when acting on a particular pixel. Through this training paradigm, we enable MLLMs to recognize the object at the category level, understand which regions can be manipulated and which cannot at the region level, and ultimately generate precise coordinates and directions for manipulation at the pose level.

During inference, we employ the chain-of-thought ~\cite{wei2022chain} flow, consistent with the training flow, to make the model's predictions more interpretable. This allows us to understand the thought process of the model in obtaining the final pose prediction. The final prediction is depicted in Fig. ~\ref{fig:teasor}. Given an RGB image featuring an object and a text prompt, our method generates the contact pixel coordinate on the 2D image and an end-effector direction. Additionally, depth information projects the pixel coordinate into 3D space. 
After the initial contact is established, we design an active impedance adaptation policy to determine the movement by forecasting the upcoming waypoints in a close-loop manner. 
Specifically, this module applies small forces in the surrounding directions based on the current pose. It aims to identify the direction that yields the maximum movement, which is then chosen as the next pose. This method relies on force feedback generated along the axes and the object to adaptively adjust the direction and predict the trajectory.


In real-world testing, we observe challenges that may diverge from the simulated learning environment. For instance, in the real world, manipulating a door with a handle using a short suction gripper might require placing the end-effector at a distance from the handle to prevent collisions, which is different from the simulator. 
To address these variations, we draw inspiration from test-time adaptation (TTA)~\cite{yang2023exploring,liu2023vida}. 
TTA involves adjusting partial model parameters during inference based on the current test sample, enhancing the model's performance for specific real-world scenarios.
Subsequently, We design a TTA strategy tailored for robot manipulation, aiming to refine the model's understanding of real-world configurations.
Specifically, with the current test sample, we utilize the outcome of manipulation success or failure to supervise the model's assessment of whether the predicted pose can result in a successful manipulation and only update partial parameters. This allows the model to retain its original capabilities and adapt to the target domain by distinguishing between effective and ineffective poses in the target domain.
Since the model has learned to predict poses that are more likely to result in successful manipulations, when facing upcoming samples, the model tends to predict effective poses, thus enhancing the performance under specific real-world configurations.


Benefiting from the MLLMs and the designed paradigm, our approach exhibits generalization and common-sense reasoning ability in manipulation. Experiments show that in the simulator, our method achieves a promising manipulation success rate across 30 categories. Meanwhile, in real-world experiments, our method shows strong generalization ability, with or without TTA strategy. More real-world videos are shown in the supplement.

In summary, our contributions are as follows:
\begin{itemize}
\item We innovatively present a simple yet effective approach that transforms the ability of MLLMs into object-centric robot manipulation.
\item We design a chain-of-thought fine-tuning and inference strategy that exploits MLLMs' reasoning ability to enable robust and explainable end-effector's pose predictions.
\item Experiments across extensive categories demonstrate the generalization ability of ManipLLM.
\end{itemize}

\section{Related Works}

\begin{figure*}[t]
\begin{center}
    \includegraphics[width=\linewidth , height=8cm]{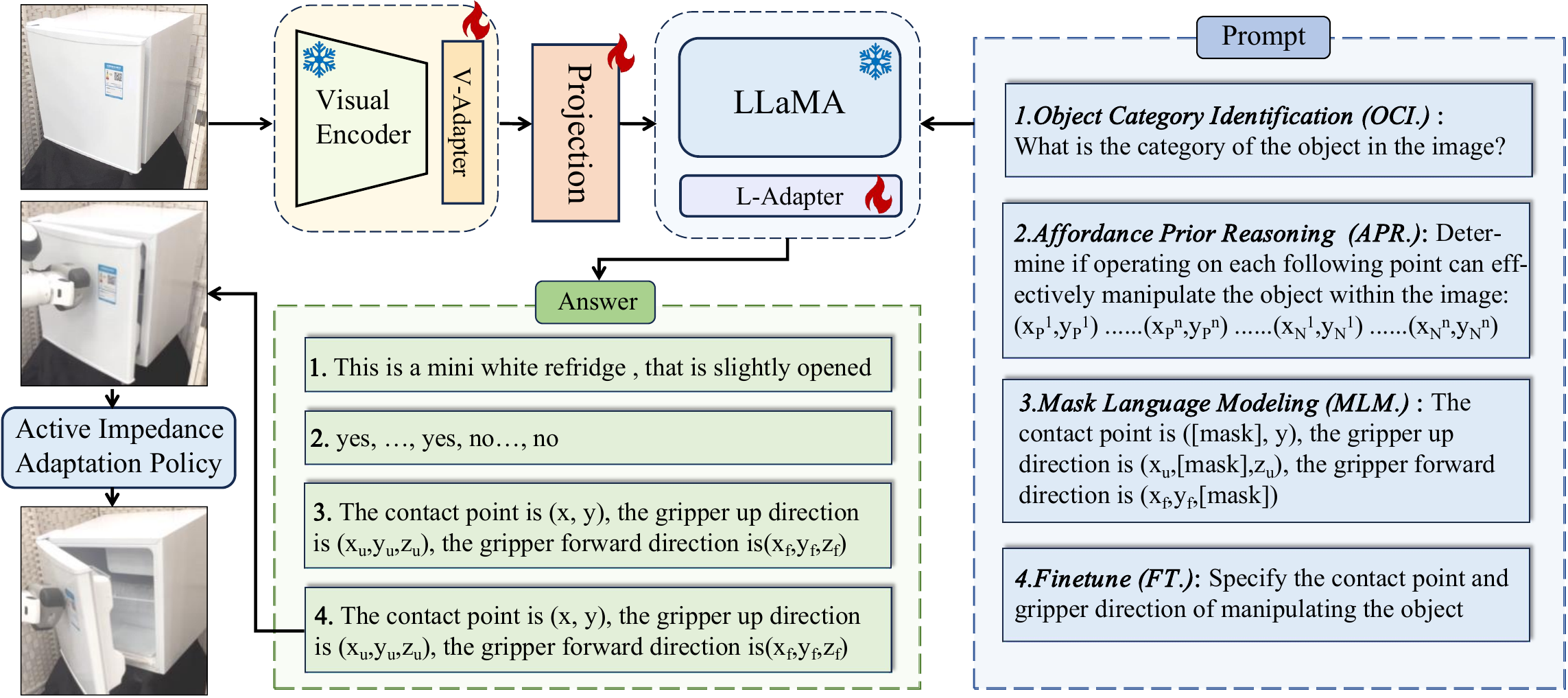}
\end{center}
\caption{Training details of ManipLLM. This paradigm contains four training tasks, enabling the model to recognize the current object (category-level), understand which regions can be manipulated (region-level), and finally generate a precise end-effector pose (pose-level).
}
\label{fig:method}
\vspace{-0.4cm}
\end{figure*}

\subsection{Robotic Manipulation}
Robotic manipulation has emerged as a pivotal research domain due to its extensive applicability.
One widely used approach is state-based reinforcement learning (RL)~\cite{joshi2020robotic,andrychowicz2020learning,yarats2021mastering,geng2022end}. 
Some works have identified the possibility of using the pure state as the policy input~\cite{andrychowicz2020learning}. However, when it comes to more complex settings, vision-based observation~\cite{Mo_2019_CVPR,xu2022universal,eisner2022flowbot3d,wu2021vat,huang2023voxposer,zitkovich2023rt, xu2023unidexgrasp, wan2023unidexgrasp++, gong2023arnold, yang2023equivact, wang2023sparsedff,geng2023sage} becomes necessary to perceive the environment and understand the complex scene and objects~\cite{deng2023banana, lei2023nap}.
Where2Act~\cite{Mo_2019_CVPR} proposes novel networks to predict actionable pixels and movable regions in objects, enabling meaningful interaction in various environments. Flowbot3d~\cite{eisner2022flowbot3d} also explores a vision-based method for perceiving and manipulating 3D articulated objects through predicting point-wise motion flow. Furthermore, VoxPoser~\cite{huang2023voxposer} synthesizes adaptable robot trajectories through 3D value maps derived from large language models, based on natural language instructions. RT2~\cite{zitkovich2023rt}, which transfers information to actions, holds promise for adapting more rapidly to novel situations.

However, although these methods achieve noteworthy accomplishments, they formulate the task as a black-box prediction, decreasing its interpretability. This becomes extremely severe when confronted with extensive categories of objects.
To reconcile such, ManipLLM harnesses the common sense knowledge and reasoning ability embedded within MLLMs to strengthen robot manipulation performance. We design intricate finetuning and inference strategies to enable interpretable object-centric pose prediction.

\subsection{Multimodal Large Language Models}
Extensive language models, \emph{i.e.}, LLaMa~\cite{touvron2023llama}, GPT3~\cite{floridi2020gpt} exhibit proficiency in a variety of language tasks given their powerful reasoning ability.
Building upon these, Multimodal Large Language Models~\cite{li2022blip,alayrac2022flamingo,liu2023visual,zhang2023llama, you2023make} are introduced to bridge RGB visual images and text.
The representative LLaMa-Adapter~\cite{zhang2023llama} generalizes to image conditions for multi-modal reasoning, achieving competitive results in both vision and multi-modal tasks. 

However, despite the considerable achievements of MLLMs, their object-centric manipulation ability is still under-explored.
Aiming to bridge this gap, our work pioneers in injecting manipulation capabilities into existing MLLMs while preserving their original reasoning ability. By doing so, the finetuned model not only possesses precise manipulation ability but is also capable of dealing with diverse category objects under interpretable thinking.



\section{Method}




\subsection{Fine-tuning Strategy}
In this section, we demonstrate how we empower MLLMs with manipulation capabilities. As shown in Fig. \ref{fig:method}, we design fine-tuning tasks at the category level, region level, and pose level, allowing the model to progressively and reasonably predict poses for object-centric robot manipulation.

\subsubsection{Model Architecture} 
We adopt the MLLM, LLaMa-Adapter~\cite{zhang2023llama}, as our backbone and follow its training strategy. Given an RGB image $I\in \mathbb{R}^{H\times W \times 3}$, we adopt the visual encoder of CLIP~\cite{radford2021learning} to extract its visual feature. 
While text prompts $T$ are encoded into a text feature using the tokenizer of the pre-trained LLaMa~\cite{touvron2023llama}.
After aligning visual and text feature representation with the multi-modal projection module, LLaMa is required to conduct multi-modal understanding and give correct answers. During training, we only fine-tune the injected adapters~\cite{hu2021lora} in visual CLIP and LLaMa~\cite{touvron2023llama}, along with the multi-modal projection module, while freezing the major parameters. This aims to reserve the powerful abilities of existing MLLMs and further empower the model with capabilities in manipulation. 

\subsubsection{Fine-tuning Tasks Formulation}
\label{sec:ft}
We design a training paradigm to fine-tune the MLLM and stimulate the model to generate interpretable pose predictions for object-centric manipulation.

\textit{Object Category Identification (OCI.):}
To successfully manipulate the object, the model needs to understand the category of the object it is facing, as objects of the same category share common geometric properties. As illustrated in the first prompt in Fig. \ref{fig:method}, we formulate the prompt as ``What is the category of the object in the image?".
It's worth mentioning that the MLLMs have been trained on a diverse set of objects in the real world, making them highly capable of category identification and generalization. In contrast, the object categories in the simulator are very limited, with a maximum of 30 to 50~\cite{mo2019partnet}. Updating the learning process in the simulator might lead to a loss of MLLMs's powerful object category identification ability and robust generalization capability. Therefore, we do not update the model in this stage, and the goal instead is to provide a prior of category cognition for subsequent tasks, helping them extract category-specific manipulation features.

\begin{figure}[t]
\begin{center}
    \includegraphics[width=\linewidth , height=5cm]{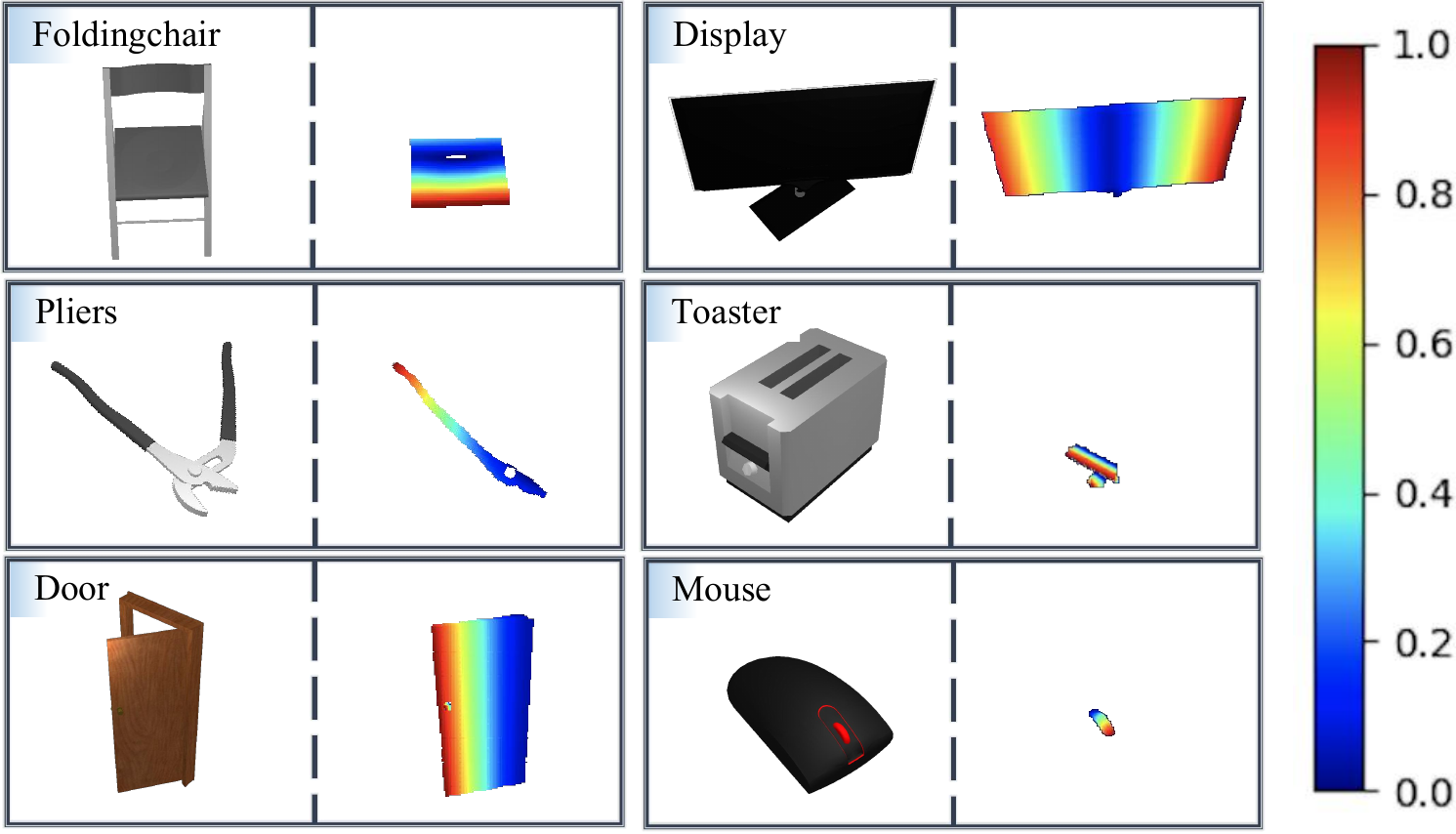}
\end{center}
\caption{Affordance map for movable parts on objects. It indicates the probability of actionability on the pixel level.
}
\label{fig:aff}
\vspace{-0.4cm}
\end{figure}

\textit{Affordance Prior Reasoning (APR.):}
This stage aims to enable the model aware where of the object region can be manipulated.
Affordance map considers the object geometric and indicates the probability of getting a moving distance if operating on certain pixels, reflecting where can act to manipulate the object. 
It can serve as a region-level affordance prior to enabling the model to have manipulation-aware localization ability. 
Inspired by Flowbot3D~\cite{eisner2022flowbot3d}, we divide the action type of the object part into ``REVOLUTE'' and ``PRISMATIC'', and collect the affordance map in the simulator accordingly. 
For the revolute part, we first find the axis of the movable object part and then enable a movement of this part along the axis. We obtain affordance map $\mathcal{A}\in\mathbb{R}^{H\times W}$ following Eq. \ref{eq:aff}:
\begin{equation}
\mathcal{A}= \frac{\mathcal{D}}{\left|\text{max}(\mathcal{D}) - \text{min}(\mathcal{D})\right|}
\label{eq:aff}
\end{equation}
The distance map, denoted as $\mathcal{D}\in\mathbb{R}^{H\times W}$, calculates the Euclidean distance of 3D positions (corresponding to each pixels) before and after the movement. 
Through applying a normalization operation based on the maximum and minimum value in distance map $\mathcal{D}$, we obtain the affordance map $\mathcal{A}\in[0,1]$, indicating the probability of actionability on pixel-level.
For the prismatic part, \emph{i.e.}, drawer, operating all points on the surface of the movable part can promote a movement. Therefore, the probability on the affordance map of the prismatic movable part are all equal to 1.
We visualize affordance maps in Fig. \ref{fig:aff}. 
For revolute parts, the affordance map reflects the regions where manipulation is possible, \emph{i.e.}, regions away from the axis.

After we obtain the affordance map, our goal is to enable the model to learn from such manipulation prior. 
Since we only have a language decoder (LLaMa) instead of a visual decoder, the model is not able to generate an affordance map directly. Therefore, we aim to translate the visually represented affordance map to linguistic affordance prior. Specifically, we randomly select $n$ positive pixels with an affordance score higher than 0.8 and select $n$ negative pixels with an affordance score lower than 0.2 as training samples. The negative samples cover both the pixels on parts that cannot be moved and the pixels on parts that can be moved but have a low affordance score, \emph{i.e.}, pixels close to the revolute axis. As shown in the second prompt in Fig. \ref{fig:method}, we formulate the text prompt with the coordinates of the selected pixels as ``Determine if operating on each following point can effectively manipulate the object within the image: ($x_{P}^{1}$,$y_{P}^{1}$)..($x_{P}^{n}$,$y_{P}^{n}$)..($x_{N}^{1}$,$y_{N}^{1}$)..($x_{N}^{n}$,$y_{N}^{n}$)'', where $P$ and $N$ denote positive and negative samples. The corresponding ground truth answer is formulated as``yes, yes .... no, no...'' with $n$ ``yes'' and $n$ ``no'' based on affordance scores. This is supervised under cross-entropy loss $\mathcal{L}_{A}$, enabling the model aware where of the object region can be manipulated and facilitating the model latter predict contact position that can promote a movement.

\textit{Finetuning (FT.) and Mask Language Modeling (MLM.):}
These tasks aim to enable the model to generate the precise end-effector pose.
In the simulator, when pre-collecting training data, if the manipulation is successful, we record the RGB image and the corresponding end-effector pose, which are used as model input and answer ground truth.
For task finetuning (FT.), as shown in the last prompt in Fig. \ref{fig:method}, we design the input text prompt for pose prediction as ``Specify the contact point and gripper direction of manipulating the object." The answer is formulated as ``The contact point is ($x$, $y$), the gripper up direction is ($x_{u}$, $y_{u}$, $z_{u}$), and the gripper forward direction is ($x_{f}$, $y_{f}$, $z_{f}$)". To decrease the difficulty of direction regression prediction, we transform it to classification prediction by discretizing the continuous numbers in the normalized direction vectors into 100 discrete bins [-50,50], with each bin spanning 0.02. The output is supervised under cross-entropy loss $\mathcal{L}_{F}$.

However, we found that directly fine-tuning the model for pose prediction leads to inaccuracies. Therefore, to facilitate the prediction of pose, in task Masked Language Modeling(MLM), we mask out the value of coordinate or direction vectors in the input text prompt and promote the model to infill the missing characters, as shown in the third prompt in Fig. \ref{fig:method}. This is supervised by the unmasked answer under cross-entropy loss $\mathcal{L}_{M}$ to stimulate the model's ability in pose prediction. The model learns to predict reasonable contact positions benefit from affordance prior learning. As for predicting the appropriate direction, we observe that MLLMs inherently possess direction awareness, such as being able to reason out ``pull the door toward you''. The training maps such direction cognitive descriptions and direction vectors to a consistent representation, enabling the prediction of the end-effector direction.

\begin{figure}[t]
\begin{center}
    \includegraphics[width=\linewidth , height=6.5cm]{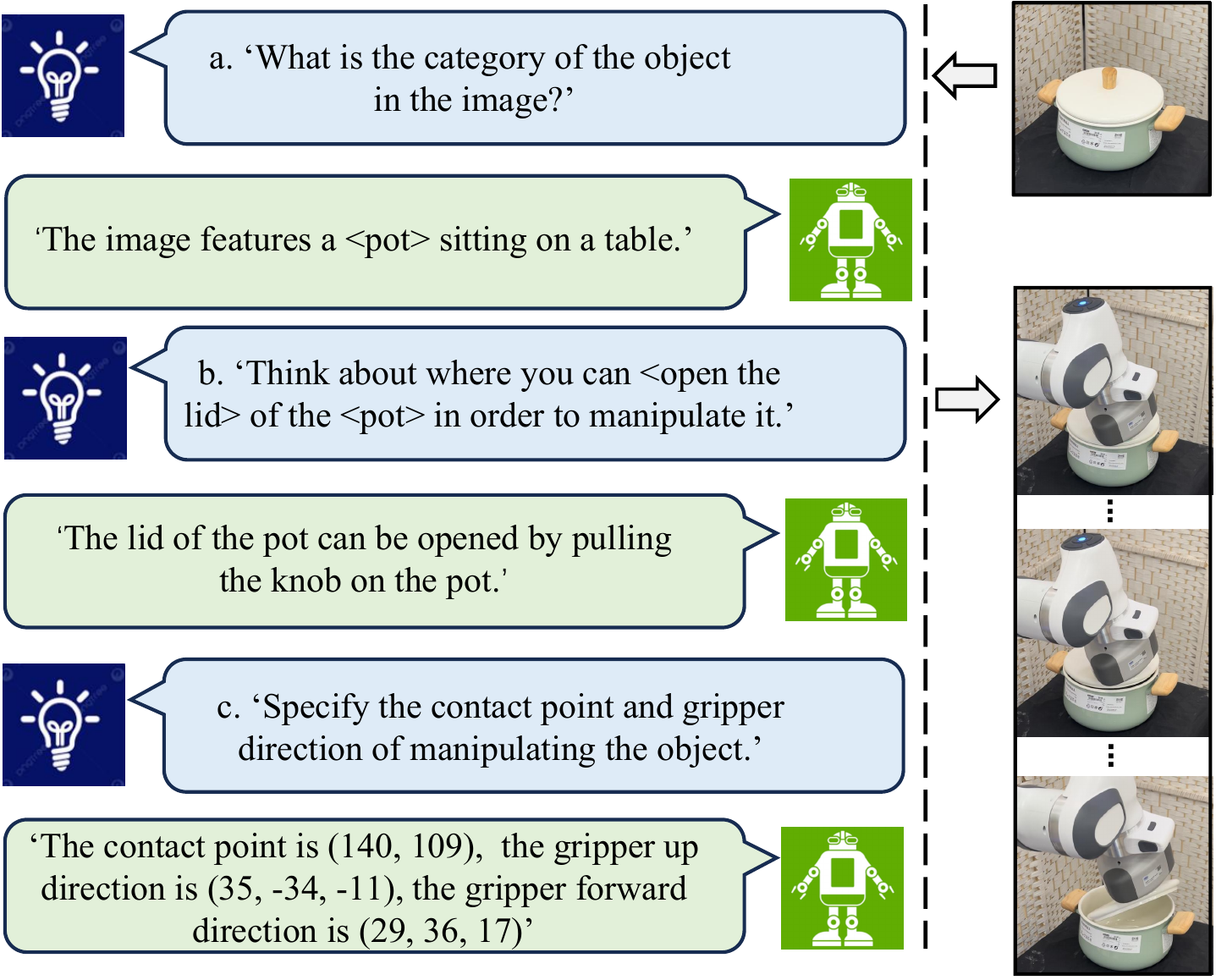}
\end{center}
\caption{The chain-of-thought inference process of ManipLLM.
}
\label{fig:infer}
\vspace{-0.4cm}
\end{figure}

\textit{Training and Inference.}
During training, the aforementioned tasks are trained simultaneously under the total objective function: $\mathcal{L}$ = $\mathcal{L}_{A}$+$\mathcal{L}_{M}$+$\mathcal{L}_{F}$. During inference, we adopt chain-of-thought reasoning to simulate the model to generate a precise initial contact end-effector pose interpretively. As shown in Fig. \ref{fig:infer}, the reasoning process follows the three steps that are consistent with the training tasks.
The model finally outputs pixel coordinate $(x, y)$, gripper up direction $(x_u, y_u, z_u)$, and gripper forward direction $(x_f, y_f, z_f)$. We utilize the depth map $\mathbb{D}^{H\times W}$ to project the contact point into the 3D manipulation space $(x, y, z)$. The gripper up direction and gripper forward direction jointly formulate the end effector's rotation.
Together with the predicted direction, they jointly determine the pose of the end-effector to establish the initial interaction with the object.

\subsection{Active Impedance Adaptation Policy}
\label{sec:aia}
After the initial interaction with the object is established, we apply a close-loop heuristic policy to adaptively generate upcoming waypoints under impedance control, \emph{i.e.} the trajectory for opening the door. 
In the task of manipulating articulated objects, where we have limited freedom to move things around, it can be quite tough to figure out the best way to do it.
For example, when trying to open a door, the best way to do it often involves moving it in a very specific direction along the axis of the door frame. 
To deal with these difficulties, the proposed policy aims to adjust how we interact with things based on impedance force feedback, which can handle different scenarios effectively. 
In contrast with leveraging a model to predict each following pose, such a heuristic policy is much more efficient.

This policy is employed within a loop to incrementally accomplish a manipulation task. At each iteration of the loop, it adaptively predicts the current direction that is best suited for the current state of the object, based on the previous step's forward direction.
We use the initial iteration as an example.
Given the predicted forward direction $d_i=(x_{f},y_{f},z_{f}), i=1$, a random perturbation $\zeta$ is introduced such that $\|\zeta\|<\epsilon_1$, where $\epsilon_1$ denotes a small positive constant. This procedure is reiterated $N$ times for $d_i$, resulting in a set of directions represented as $D=\{d_{ij}=d_i+\zeta_j\}_{j\in\{0,1,2,\ldots,N\}}$, where $d_{i0}$ represents $d_i$, $\zeta_0=\mathbf{0}$.
Utilizing impedance control, a force ${f_j}$ (${f_j}$ has a direction defined by $d_{ij}$ and $\|f_j\|=\epsilon_2$, where $\epsilon_2$ denotes another small positive constant) is applied to each direction $d_{ij}$ within $D$. The optimal direction $d_{opt}$ is then determined based on the observed end-effector movements $\delta_j$. 
We assume that in constrained object manipulation tasks, greater movements represent the efficacy of the applied force direction. Thus, the best forward direction is generated as the following to determine the current end-effector's pose:

$$ d_{\mathrm{opt}}, \ \mathrm{opt} = \underset{j \in \{0,1,\ldots,N\}}{\arg\max} \ \| \delta_j \| $$

By doing so, we determine the optimal movement pose given the current object state by considering the force feedback along the axis and ensuring a smooth trajectory.

\subsection{Sim-to-real Transfer}

Though the model is trained to perform well in the simulator, in the real world, robots often encounter more unique situations, \emph{i.e.}, environment, or hard device configuration, which may differ significantly from what simulators simulate, leading to a sim-to-real gap. 
To bridge this gap, we design a Test-Time Adaptation (TTA) strategy tailored for manipulation. TTA, as described by ~\cite{liu2023vida,yang2023exploring}, involves updating partial model parameters during inference based on the current test sample, enhancing the model's performance for specific real-world scenarios.
To determine which parameters to update for pose prediction during TTA, we analyze the outcomes of the reasoning steps during inference in Fig.~\ref{fig:infer}.
We observe that the reasoning abilities of ManipLLM, benefit from LLaMa~\cite{touvron2023llama}, continue to exhibit strong performance in real-world scenarios. It can accurately recognize objects depicted in images and comprehend how to manipulate them. Its orientation awareness is also robust, ensuring the robustness of ManipLLM's direction predictions. 
Even though there might be imprecise directions, with the active impedance adaptation policy introduced in Sec.~\ref{sec:aia}, we can adjust the direction to a more optimal state.
In contrast, position predictions are susceptible to domain gaps caused by factors like lighting and texture. Consequently, we adjust the visual perception for the target domain during TTA by only updating the V-Adapter in Fig.~\ref{fig:method}.

Specifically, given the current test sample, we introduce an additional reasoning step to prompt the model to assess whether the predicted position can lead to a successful manipulation. The text prompt used in this step is consistent with the training phase of ``Affordance Prior Reasoning'', which is ``Determine if operating on the following point can effectively manipulate the object within the image: (x, y)." 
The contact position predicted by the model is the region that they believe can lead to a successful manipulation. Therefore, the responses to this question are consistently ``yes."
We obtain the ground-truth result based on whether, in the real world, the object was successfully manipulated, forming either a ``yes" or ``no" as the supervision signal to supervise the previous answer. 
By implementing this process, we enable the model to distinguish between effective and ineffective predictions in the target domain. This adjustment allows the model to predict valid poses when facing subsequent test samples, thereby adapting to the specific real-world configuration.

\section{Experiment Results}

\subsection{Training Details}
\textbf{Data Collection.}
We adopt SAPIEN~\cite{Xiang_2020_SAPIEN} and the PartNet-Mobility dataset to set up an interactive environment for our task, with VulkanRenderer of high-efficiency rasterization-based renderer.
We use a Franka Panda Robot with flying suction gripper as the robot actuator.
We sample the training data offline with approximately 10,000 manipulation success samples across 20 categories. 
We randomly select a contact point $p$ on the movable part and use the opposite direction of its normal vector as the end-effector orientation for interacting with the object. If successful manipulation is achieved, we record it as a successful sample. 
The tasks involve several pulling action primitives, where the movement direction of the object part and the opposite of end effector direction are within the same hemisphere, i.e., open the drawer, open the door, rotate the pliers, lift the lid, etc.

\textbf{Training Details.}
We finetuned LLaMA-Adapter~\cite{zhang2023llama} on a 40G A100 GPU for 10 epochs, with an epoch costs around an hour. 
It includes pre-trained CLIP ~\cite{radford2021learning} as the visual encoder, 7B LLaMA~\cite{touvron2023llama} model as the decoder, and multi-modal projection module of 32 transformer layers. The n in Affordance Prior Reasoning is set to 20.

\textbf{Evaluation Metric.}
We adopt the manipulation success rate to reflect the outcome of the manipulation which is the ratio of the number of successfully manipulated samples divided by the total number of all test samples. As for the definition of success sample, we adopt the binary success definition which is measured by thresholding the move distance of the object part at $\delta$: success = $1(\delta_{dis}>\delta)$. We set $\delta = 0.01$ or $\delta = 0.1$ for initial movement or long-distance movement, respectively, meaning that the gap between the start and end part 1-DoF pose is greater than 0.01 or 0.1 unit length. Initial movement is a prerequisite for long-distance movement and can effectively reflect the model's ability to predict end-effector pose. Both initial and long-distance movements apply active impedance adaptation policy to adjust movement direction.

\begin{table*}[tb]
	\begin{center}
	
	\small
	    \setlength{\tabcolsep}{1.5mm}{
		\begin{tabular}{c| cc c c c c c c c c c c c c c c}
  \hline
	\multirow{2}{*}{\textbf{}}&\multirow{2}{*}{\textbf{}} &\multicolumn{15}{c}{\textbf {Train Categories}}\\
 Method
    & \includegraphics[width=0.035\linewidth]{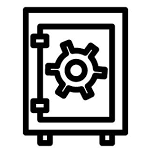}
    &\includegraphics[width=0.035\linewidth]{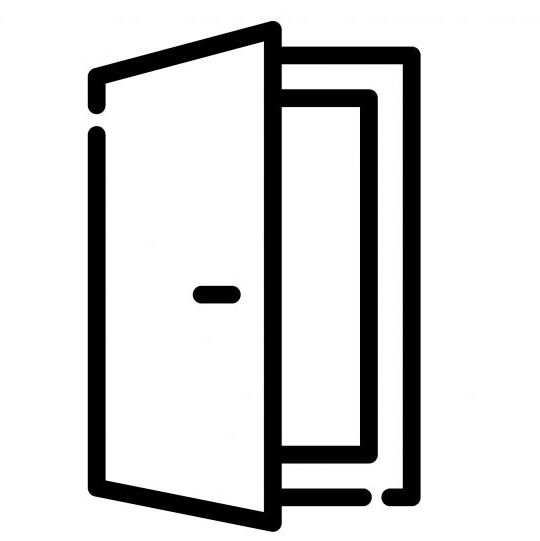}
        &\includegraphics[width=0.035\linewidth]{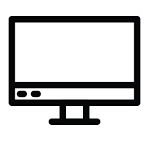}
        &\includegraphics[width=0.035\linewidth]{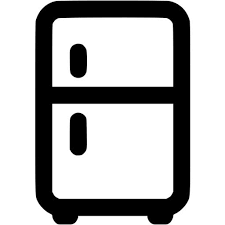}
        &\includegraphics[width=0.035\linewidth]
        {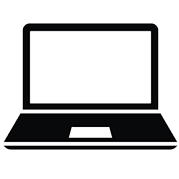}
        &\includegraphics[width=0.035\linewidth]
        {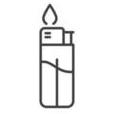}
        &\includegraphics[width=0.035\linewidth]{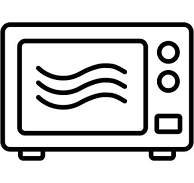}
        &\includegraphics[width=0.035\linewidth]{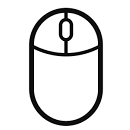}
        &\includegraphics[width=0.035\linewidth]{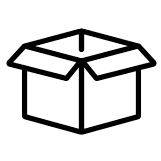}
        &\includegraphics[width=0.035\linewidth]{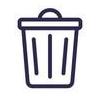}
        &\includegraphics[width=0.035\linewidth]{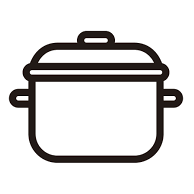}
        &\includegraphics[width=0.035\linewidth]{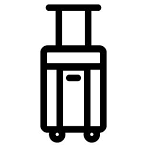}
        &\includegraphics[width=0.035\linewidth]{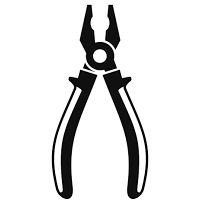}
        &\includegraphics[width=0.035\linewidth]{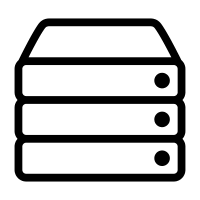}
        &\includegraphics[width=0.035\linewidth]{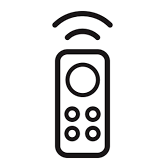}
        &\includegraphics[width=0.035\linewidth]{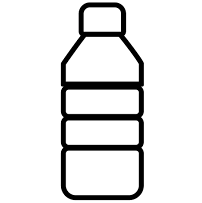}\\\hline\hline
        
        Where2Act~\cite{Mo_2019_CVPR}& 0.26&	0.36&	0.19&	0.27&	0.23&	0.11&	0.15&	0.47&	0.14&	0.24&	0.13&	0.12&	0.56&	0.68&	0.07&	0.40\\ 
  UMPNet~\cite{xu2022universal}&  0.46&	0.43&	0.15&	0.28&	\textbf{0.54}&	0.32&	0.28&	0.56&	0.44&	0.40&0.10&	0.23&	0.18&	0.54&	0.20&	0.42 \\
  FlowBot3D~\cite{eisner2022flowbot3d}&  0.67&	0.55&	0.20&	0.32&	0.27	&0.31&	0.61&	\textbf{0.68}&	0.15&	0.28&	0.36&	0.18&	0.21&	0.70&	0.18&	0.26\\
  Implicit3D~\cite{zhong20233d}&  0.53& 	0.58&	0.35&	0.55&	0.28&	\textbf{0.66}&	0.58&	0.51&	0.52&	0.57&	0.45&	0.34	&0.41&	0.54&	0.39&	0.43\\\hline

  Ours& \textbf{0.68}&	\textbf{0.64}	&\textbf{0.36}&	\textbf{0.77}&	0.43&	0.62&	\textbf{0.65}&	0.61&	\textbf{0.65}&	\textbf{0.52}&	\textbf{0.53}&	\textbf{0.40}&	\textbf{0.64}&	\textbf{0.71}&	\textbf{0.60}&	\textbf{0.64} \\ \hline 
  Ours (long)&0.68	&0.62&	0.28&	0.76	&0.43&	0.62&	0.65&	0.61	&0.61	&0.45&	0.43&	0.38&	0.62&	0.71&	0.60&	0.63 \\\hline
			\hline
	\multirow{2}{*}{\textbf{}}&\multirow{2}{*}{\textbf{}}&\multicolumn{4}{c|}{\textbf {Train Categories} }&

			\multicolumn{10}{c}{\textbf {Test Categories}}\\
			
	Method
    & \includegraphics[width=0.035\linewidth]{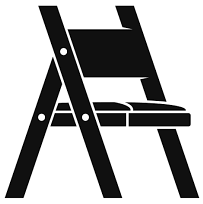}
        &\includegraphics[width=0.035\linewidth]{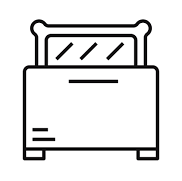}
        &\includegraphics[width=0.035\linewidth]{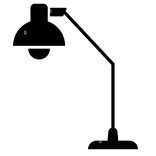}
        &\includegraphics[width=0.035\linewidth]{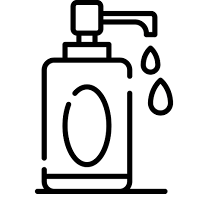}
        &\multicolumn{1}{c|}{{\textbf {AVG}} }
        &\includegraphics[width=0.035\linewidth]{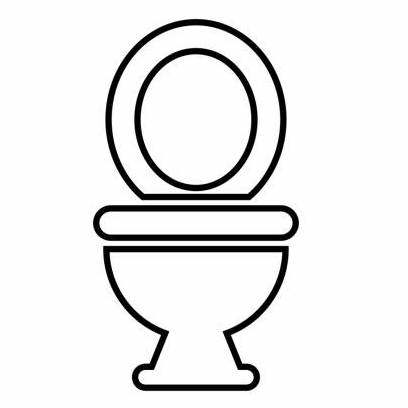}
        &\includegraphics[width=0.035\linewidth]{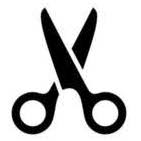}
        &\includegraphics[width=0.035\linewidth]{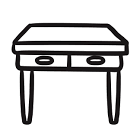}
        &\includegraphics[width=0.035\linewidth]{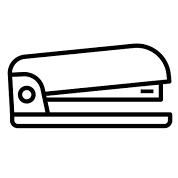}
        &\includegraphics[width=0.035\linewidth]{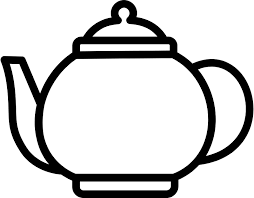}
        &\includegraphics[width=0.035\linewidth]{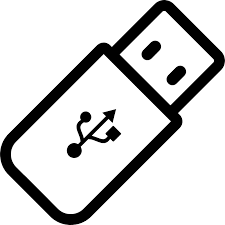}
        &\includegraphics[width=0.035\linewidth]{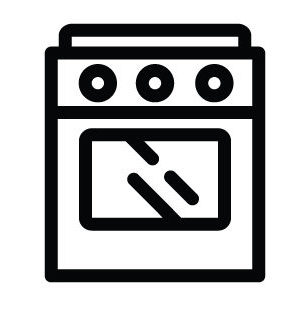}
        &\includegraphics[width=0.035\linewidth]{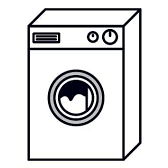}
        &\includegraphics[width=0.035\linewidth]{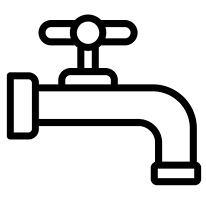}
        &\includegraphics[width=0.035\linewidth]{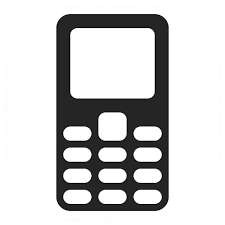}
        &{\textbf {AVG}}\\\hline\hline
  Where2Act~\cite{Mo_2019_CVPR}& 0.13&	0.18	&0.13&	0.40&	\multicolumn{1}{c|}{0.26}& 0.18&	\textbf{0.35}	&0.38&	0.28&	0.05&	0.21&	0.17&	0.20&	0.15	&0.15	&0.21 \\ 
  UMPNet~\cite{xu2022universal}&  0.22&	0.33&	0.26&	0.64&	\multicolumn{1}{c|}{0.35}& 0.42&	0.20&	0.35	&0.42&	0.29&	0.20&	0.26&	0.28&	0.25&	0.15&	0.28\\
  FlowBot3D~\cite{eisner2022flowbot3d}&  0.17&	0.53&	0.29	&0.42&	\multicolumn{1}{c|}{0.37}& 0.23&	0.10&	0.60&	0.39&	0.27&	0.42&	0.28&	0.51&	0.13&	0.23&	0.32
  \\
  Implicit3D~\cite{zhong20233d}&  0.27&	0.65&	0.20&	0.33&	\multicolumn{1}{c|}{0.46}& \textbf{0.45}&	0.17&	0.80&	0.53	&0.15&	0.69&	0.41&	0.31&	\textbf{0.30}&	0.31&	0.41\\\hline

  Ours& \textbf{0.41}&	\textbf{0.75}&	\textbf{0.44}&	\textbf{0.67}&	\multicolumn{1}{c|}{\textbf{0.59}}& 0.38	&0.22&	\textbf{0.81}	&\textbf{0.86}&	\textbf{0.38}&	\textbf{0.85}	&\textbf{0.42}	&\textbf{0.83}&	0.26&	\textbf{0.38}&	\textbf{0.54} \\ \hline 
  Ours (long)& 0.37&	0.75&	0.44&	0.67&	\multicolumn{1}{c|}{0.57}& 0.34&	0.22&	0.81&	0.86&	0.30&	0.85&	0.42&	0.80&	0.26&	0.38&	0.52\\\hline

		\end{tabular}}
		\vspace{-0.1cm}
	\caption{Comparisons of our method against baseline methods.} 
 \vspace{-0.1cm}
 \label{tab:pull}	
	\end{center}
\end{table*}

\subsection{Quantitative Comparison}
We compare ManipLLM against four representative baselines, including Where2Act~\cite{Mo_2019_CVPR}, UMPNet~\cite{xu2022universal}, Flowbot3D~\cite{eisner2022flowbot3d}, and Implicit3D~\cite{zhong20233d}.
For simplicity, we conduct the comparisons with other approaches only on initial movement setting since this can reflect how well the model can perform given the initial state of the object, which is the preliminary condition in realizing the whole long-distance movement. 
For a fair comparison, all methods are under the same train/test split and end-effector setting. 

\textbf{Where2Act~\cite{Mo_2019_CVPR}}: It takes point-cloud as input and estimates per-point scores, selecting the point with the highest score as contact point. It further predicts 100 end-effector orientations and selects the orientation with the highest score to formulate the contact pose. For fair comparison, we alter the used parallel gripper to suction gripper.

\textbf{UMPNet~\cite{xu2022universal}}: 
Following UMPNet, we execute manipulation on the contact point that UMPNet predicts with the orientation perpendicular to the object surface.

\textbf{Flowbot3D~\cite{eisner2022flowbot3d}}: 
It predicts motion direction on the point cloud, denoting it as `flow'. The point with the largest flow magnitude serves as the interaction point, while the direction of the flow represents end-effector's orientation.

\textbf{Implicit3D~\cite{zhong20233d}}:
It develops a manipulation policy for downstream tasks that utilizes the Transporter to detect keypoints for 3D articulated objects. The keypoints are then used to determine end-effector pose.

Our current experimental settings involve training on a wider range of object categories. Consequently, this poses challenges in extracting features and learning characteristics from these wide categories, which may lead to a decrease in manipulation success rate compared to the original papers. However, our method, by retaining the common-sense reasoning capabilities of MLLMs and injecting manipulation abilities, ensures strong generalization across diverse categories.
It's worth noting that other methods also require a movable mask during testing, whereas our method achieves this \textbf{without the need for movable part mask}.

\textbf{Voxposer~\cite{huang2023voxposer}}:
We also compare our method with VoxPoser for the ``Opening Drawer'' task. The success rate of VoxPoser is $14.0\%$ while ours is $69.0\%$. We found that VoxPoser struggles to find the correct grasping pose and a suitable moving trajectory for challenging cases.

\begin{figure*}[t]
\begin{center}
    \includegraphics[width=\linewidth , height=5.5cm]{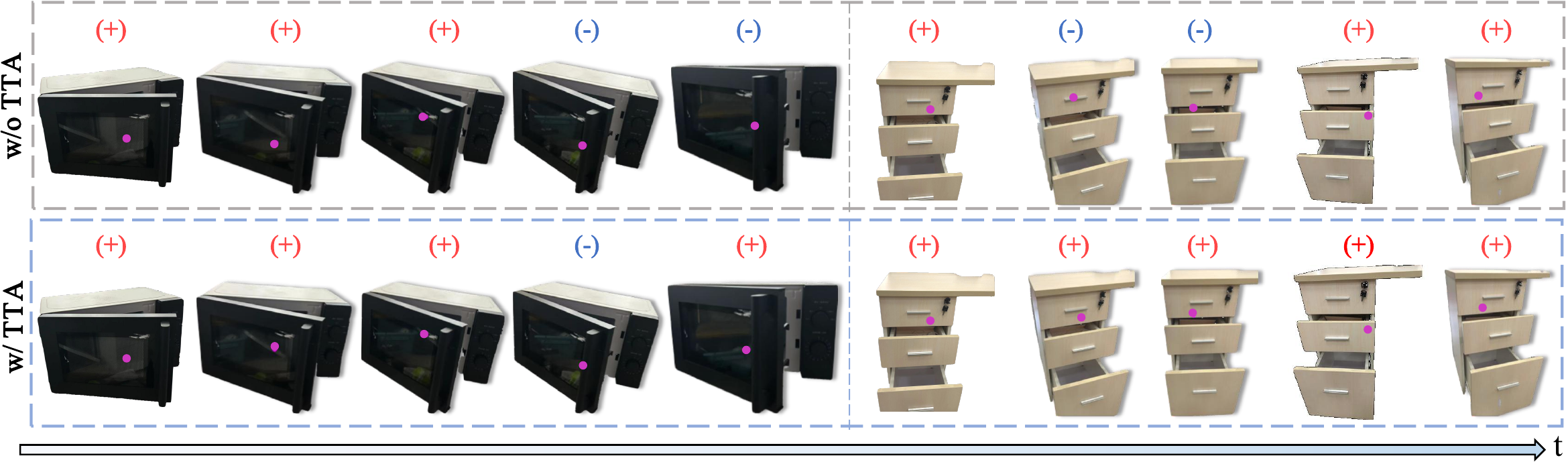}
\end{center}
 \vspace{-0.4cm}
\caption{Visualizations of TTA process in real-world scenarios. The center of pink dot represents the predicted contact position.
}
\label{fig:vis}
\end{figure*}

\subsection{Ablation and Analysis}
To elucidate the contribution and effectiveness of individual modules within our approach, we conduct extensive ablation studies. The results are measured on novel instances in train categories with initial movement.

\textbf{Effectiveness of tasks in the training paradigm.}
\begin{table}
    \centering
    \begin{tabular}{cccc|cc|c}
     \toprule
    \multicolumn{4}{c}{Train}\vline& \multicolumn{2}{c}{Test}\vline& \multirow{2}{*}{\textbf{AVG} }\\ \cmidrule{1-4} \cmidrule{5-6} 
                                        {FT.} & {OCI.} & {MLM.} & {APR.}& {COT.}
                                        & {AIA.}\\ \midrule \midrule
$\checkmark$   &-    & -    & -    &$\checkmark$&$\checkmark$& 0.41   \\ 
$\checkmark$   &$\checkmark$    & -    & -    & $\checkmark$&$\checkmark$ &0.44 \\ 
$\checkmark$   &$\checkmark$    & $\checkmark$  & -    & $\checkmark$&$\checkmark$ & 0.50   \\ 
  $\checkmark$&$\checkmark$   & $\checkmark$   & $\checkmark$&$\checkmark$  &$\checkmark$ & 0.59   \\\hline
$\checkmark$&$\checkmark$   & $\checkmark$   & $\checkmark$  &-&$\checkmark$ & 0.56\\
$\checkmark$&$\checkmark$   & $\checkmark$   & $\checkmark$  &$\checkmark$&-& 0.55\\
\hline
    \end{tabular}
  \caption{Ablation analysis of each training task in the training paradigm and strategies in inference.}
  \label{tab:abla}
  \vspace{-0.4cm}   
\end{table}
 In Table.~\ref{tab:abla}, we gradually add each task in the training paradigm to show the effectiveness of each.  
\textit{Finetuning(FT.):} In the first row of Table~\ref{tab:abla} with only fine-tuning, the last prompt in Fig.~\ref{fig:method}, we found that introducing this single task allows the model to have some manipulation capability, showcasing the strong learning ability of the large model.
\textit{Object Category Identification(OCI.):} Subsequently, in the second row of Table~\ref{tab:abla}, we introduce the task of object category identification, the first prompt in Fig.~\ref{fig:method}. It enables the model to discover commonalities in manipulating objects of the same or similar categories, thus enhancing the model's manipulation abilities by $+3\%$. 
\textit{Mask Langugae Modeling(MLM.):} Next, in the third row of Table~\ref{tab:abla}, also the third prompt in Fig.~\ref{fig:method}, we randomly mask values in the coordinates or direction vectors to force the model to predict precise pose. This task stimulates the model's ability of localization that it previously lacked and enables the model to map the direction common sense reasoning with $SO(3)$ direction representation, thus improving the performance by $+6\%$.
\textit{Affordance Prior Reasoning(APR.):}Finally, in the fourth row of Table~\ref{tab:abla}, we introduce the affordance prior reasoning task, the second prompt in Fig.~\ref{fig:method}, allowing the model to learn manipulation-aware localization and predict accurate contact positions. It thus significantly improves the manipulation success rate by $+9\%$.

\textbf{Effectiveness of strategies in inference.}
\textit{Chain-Of-Thought Reasoning (COT.):}
The COT reasoning strategy aims to enable the model to generate the final pose prediction under a transparent and reasonable thinking process. For comparison, we ask the model to generate the final pose prediction directly without the thinking process in Fig.~\ref{fig:infer}.
As shown in the second-to-last last row of Table~\ref{tab:abla} w/o COT, we find that this decreases the performance by $-3\%$ compared with applying COT during inference. This emphasizes the essential of enabling the model to predict under a transparent and interpretable process.

\textit{Active Impedance Adaptation (AIA.):}
The active impedance adaptation policy adaptively adjusts the pose to adapt the current object state under impedance control. In the last row of Table~\ref{tab:abla} w/o AIA., we employ a straightforward control policy, which operates by moving directly to the desired position under invariant direction, 
In contrast, Active Impedance Adaptation policy applies impedance control to effectively adjust the direction based on force feedback, thus enabling a smooth manipulation trajectory in the long-term movement and improving the long-distance movement by $-4\%$. This policy is particularly crucial for long-distance movement, and the performance would decrease from 0.57 to 0.50 if not applied.

\subsection{Real-world Evaluation}

We conduct experiments that involve interacting with various real-world household objects. 
We employ a Franka Emika robotic arm with cobot pump suction gripper and utilize a RealSense 415 camera to capture RGB image and depth map.
To address the sim-to-real problem: 1) during training, we utilize the LLAMA-Adapter pretrained in the real-world and employ a method that combines injection and finetuning of the adapter to make it learn new downstream tasks. This training strategy allows the model to retain its strong perception and reasoning abilities in the real-world while equipping it with the capability to perform new manipulation tasks. 2) When collecting data in the simulator, we employ domain randomization to increase scenario diversity by varying elements such as object part poses, camera view angles, and lighting, among others, in order to mitigate the potential sim-to-real gap. 
3) During testing, we design test-time adaptation tailored for manipulation tasks to help the model better adapt to the current scene's configuration. By finetuning only the adapters in visual modules, we shift its attention, enabling MLLMs to generate predictions that are better suited to the current scene. 

The results of real-world experiments are shown in Table~\ref{tab:real}. As illustrated in Fig.~\ref{fig:vis}, the devised TTA strategy effectively addresses discrepancies arising from real-world hardware configurations. 
In our specific hardware configuration, the suction gripper is unable to grasp the handle due to the non-smooth surface. Additionally, its head is relatively short, which presents a collision risk when closely interacting with the protruding handle.
The TTA process learns from both successful and unsuccessful scenarios, gradually adapting its predictions to align with the current configuration. 
Since this strategy is plug-and-play, in supplement, we further apply it in the simulator when dealing with unseen category objects to show its effectiveness.

\begin{table}[h]
    \centering
    \small
    \begin{tabular}{|c|*{6}{p{0.37cm}|}c|}
    \hline
    Object Category & \includegraphics[width=0.37cm]{./figure/icon/micro.png} &
    \includegraphics[width=0.37cm]{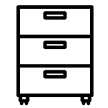}  &
    \includegraphics[width=0.37cm]{./figure/icon/trashcan.png} &
    \includegraphics[width=0.37cm]{./figure/icon/pot.png} &
    \includegraphics[width=0.37cm]{./figure/icon/fridge.png}  &
    \includegraphics[width=0.37cm]{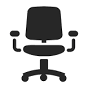} &
    \includegraphics[width=0.37cm]{./figure/icon/toilet.png}\\  
        \hline\hline
         Success/Total & 4/5 & 5/5  & 4/5& 3/5 & 4/5 & 4/5 &4/5\\
        \hline
         Distance(m) & 0.17 & 0.28  &0.10 & 0.08 & 0.14 & 0.15 & 0.18 \\
        \hline
    \end{tabular}
    \caption{Real world experiments.}
    \label{tab:real}
    \vspace{-0.4cm}   
\end{table}

\section{Conclusion}


We transform MLLMs to robotic manipulation through the chain-of-though training paradigm and equip the model with the ability to predict poses.
We introduce an active impedance adaptation policy that adjusts direction based on force feedback of the correct state to ensure a smooth moving trajectory.
ManipLLM shows strong generalization ability across extensive categories and in real-world.

\clearpage
{
    \small
    \bibliographystyle{ieeenat_fullname}
    \bibliography{main}
}


\end{document}


\maketitle

\section{Additional Experiment Details}
\subsection{More Details on Experiment Setting}
When collecting training data, to augment domain randomlization, we place the camera 4.5-5.5 units away from the object, facing the object's center, and situated in the upper hemisphere of the object at a random azimuth angle between $0^\circ$ and $360^\circ$, as well as a random altitude angle between $30^\circ$ and $60^\circ$. This boosts the variery in view angle and help to deal with view angle issue when transferring from simulator to real world.

\begin{figure}[h]

    \begin{center}
        \includegraphics[width=\linewidth]{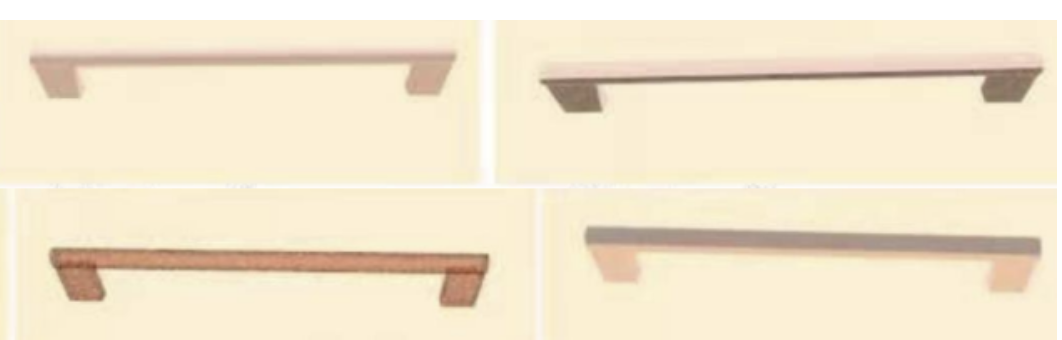}
    \end{center}
    \caption{Domain randomization on material.}
    \label{fig:mat}
\end{figure}

In simulator, we've also employed domain randomization to amplify scenario diversity, diversifying elements like lighting, materials, light position, etc,  aiming to ease sim-to-real transfer. We visualize the domain randomization of handle material in Fig.~\ref{fig:mat}.

In order to tackle the significant disparity between visual and collision shapes, we leverage the V-HACD~\cite{mamou2016volumetric}(Voxelized Hierarchical Approximate Convex Decomposition) algorithm.
This method entails voxelizing the 3D model, subsequently engaging hierarchical approximation to iteratively diminish the voxel count and amalgamate them into larger convex voxels. Subsequently, convex decomposition is applied to transform these merged convex voxels into simpler convex shapes.
\subsection{Representation for Each Category Icon}
In Table~\ref{tab:icon}, we provide an overview of the meaning of each category icon in Table 1 in the main paper. These categories, along with their corresponding objects, are sourced from PartNet-Mobility~\cite{chang2015shapenet}.
\begin{table*}[t]
    \centering
    \small
    \renewcommand{\arraystretch}{1.2}
    \setlength{\tabcolsep}{2mm}
    
    \begin{tabular}{*{10}{c}}
        \hline
        \includegraphics[width=0.5cm]{./figure/icon/safe.png} &
        \includegraphics[width=0.5cm]{./figure/icon/door.png} &
        \includegraphics[width=0.5cm]{./figure/icon/display.png} &
        \includegraphics[width=0.5cm]{./figure/icon/fridge.png} &
        \includegraphics[width=0.5cm]{./figure/icon/laptop.png} &
        \includegraphics[width=0.5cm]{./figure/icon/lighter.png} &
        \includegraphics[width=0.5cm]{./figure/icon/micro.png} &
        \includegraphics[width=0.5cm]{./figure/icon/mouse.png} &
        \includegraphics[width=0.5cm]{./figure/icon/box.png} &
        \includegraphics[width=0.5cm]{./figure/icon/trashcan.png} \\
        \hline
        Safe & Door & Display & Fridge & Laptop & Lighter & Microwave & Mouse & Box & Trashcan   \\
        \hline\hline
        \includegraphics[width=0.5cm]{./figure/icon/pot.png} &
        \includegraphics[width=0.5cm]{./figure/icon/suitcase.png} &
        \includegraphics[width=0.5cm]{./figure/icon/pliers.png}
        &\includegraphics[width=0.5cm]{./figure/icon/storage.png}
        &\includegraphics[width=0.5cm]{./figure/icon/remote.png}
        &\includegraphics[width=0.5cm]{./figure/icon/bottle.png}
        &\includegraphics[width=0.5cm]{./figure/icon/folding.png}
        &\includegraphics[width=0.5cm]{./figure/icon/toaster.png}
        &\includegraphics[width=0.5cm]{./figure/icon/lamp.png}
        &\includegraphics[width=0.5cm]{./figure/icon/dispenser.png}
        \\\hline
        Pot & Suitcase & Pliers & Storage & Remote & Bottle & Foldingchair & Toaster & Lamp & Dispenser  \\\hline\hline
        \includegraphics[width=0.5cm]{./figure/icon/toilet.png}
        &\includegraphics[width=0.5cm]{./figure/icon/scrissor.png}
        &\includegraphics[width=0.5cm]{./figure/icon/table.png}
        &\includegraphics[width=0.5cm]{./figure/icon/stapler.png} &
        \includegraphics[width=0.5cm]{./figure/icon/kettle.png}
        &\includegraphics[width=0.5cm]{./figure/icon/usb.png}
        &\includegraphics[width=0.5cm]{./figure/icon/oven.png}
        &\includegraphics[width=0.5cm]{./figure/icon/washing.png}
        &\includegraphics[width=0.5cm]{./figure/icon/faucet.png}
        &\includegraphics[width=0.5cm]{./figure/icon/phone.png} \\\hline
        Toilet & Scissors & Table & Stapler & Kettle & USB & Oven & Washingmachine & Faucet & Phone\\\hline
        
    \end{tabular}
    \caption{Representation of each category icon.}
    \label{tab:icon}	
\end{table*}

\section{More Experiments}
\subsection{Experiments for TTA in Simulator}
Because TTA (Test-Time Augmentation) is a plug-and-play strategy, we also employ this approach in the simulator when facing test categories to analyze deeper into its effectiveness. In this experiment, we utilize the success or failure of manipulations in the simulator as a supervisory signal to guide the model in determining whether the predicted pose will lead to a successful manipulation outcome. We only update the visual encoder's V-Adapter to preserve the model's inherent capabilities as much as possible while adapting to the target domain. Under this testing strategy, for the measurement of the initial movement in the test category, the success rate increases from 0.54 to 0.57, indicating an improvement with this strategy. Moreover, to maintain the model's generalization performance, the number of updated parameters is minimal. The model's capacity for updatable parameters is not extensive, resulting in a moderate increase in the success rate, showing it is still the model's intrinsic capabilities playing a more dominant role.

We further investigate and find that when statistically testing the initial 50 test samples, the manipulation success rate increases significantly, showing an improvement of approximately 0.5 compared to the same period without TTA. In subsequent tests, the rate of improvement slows down. In the final 50 test samples, the improvement is approximately 0.1 compared to the same period without TTA. We thus assume that due to the limited number of parameters in the V-Adapter, there is a finite amount of knowledge that can be learned, and the potential for performance improvement is not limitless. 

To verify this, we add adapters to more layers in the visual encoder. Our approach (in main paper) involves adding adapters only to the linear layers in the Clip encoder. In the comparative experiment, adapters are added to the transformer layers as well, increasing the number of learnable parameters more than ten times. In this scenario, the total manipulation rate remains comparable to the test without TTA (0.54). In the initial 50 test samples, the increased number of learnable parameters quickly improves the model's performance in the target domain by 0.7. However, in subsequent stages, the model's performance even lags behind the test strategy without TTA. This indicates that allowing more model parameters to adapt to the target domain may result in a loss of the model's original generalization. Therefore, we come to the conclusion that there is a trade-off between the size of learnable parameter and manipulation performance.

\subsection{Quantitative Results in Simulator}
\begin{figure}[th]
\begin{center}
    \includegraphics[width=\linewidth , height=6.3cm]{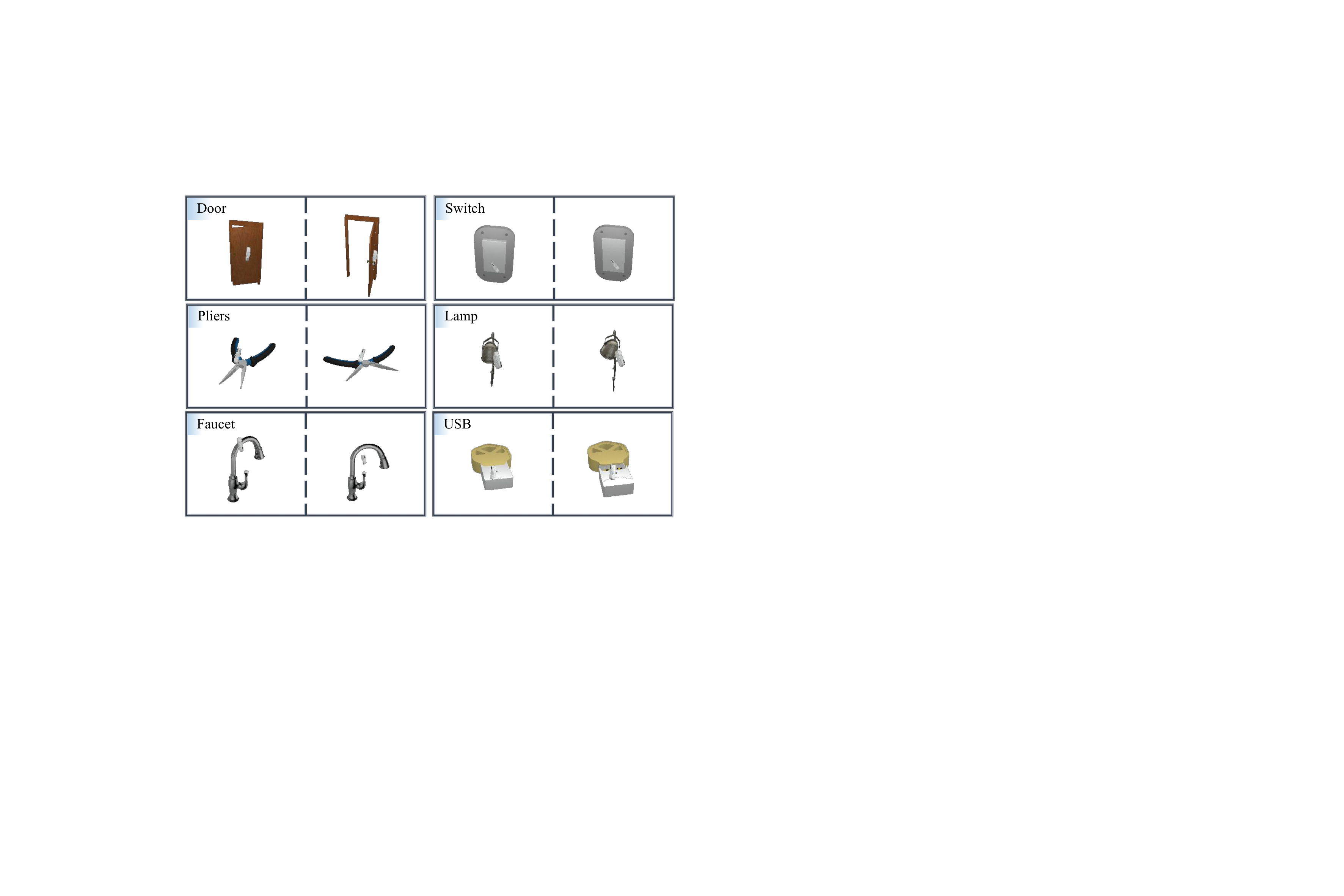}
\end{center}
\caption{Manipulation demonstration in simulator.
}
\label{fig:viz}
\end{figure}

In Fig.~\ref{fig:viz}, we visualize the initial and final object state to demonstrate how does the robot manipulate in the simulator. The distinction becomes more apparent when zoomed in at a factor of four.

\section{Real World Experiments}
\begin{figure}[t]
\begin{center}
    \includegraphics[width=\linewidth , height=6.9cm]{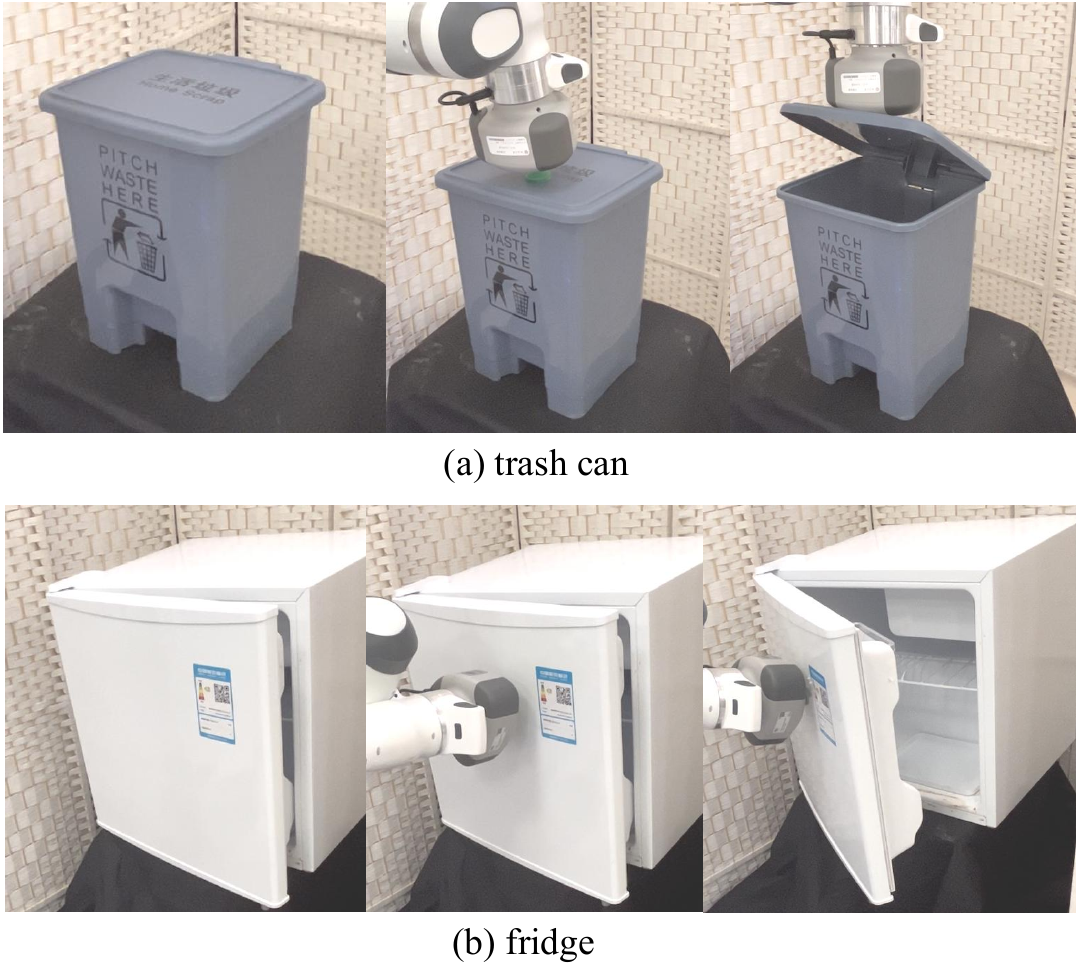}
\end{center}
\caption{Manipulation demonstration in real-world.
}
\label{fig:real}
\end{figure}
In this section, we analysis the limitation and failure cases that in our real-world setting. We observe that the primary limitation still lies in the potential for the suction cup to collide with the object's surface, especially if its orientation is not adjusted appropriately. Additionally, there is a possibility that the suction cup may fail to hold the object, as it requires a specific pressure between the cup and the object to establish a vacuum and effectively hold the item in place. Video demonstrations are shown in the \textbf{supplementary video}. In Fig.~\ref{fig:real}, we present snapshots of partial real-world experiments, illustrating the initial object state, initial contact state, and final contact state, respectively.
{
    \small
    \bibliographystyle{ieeenat_fullname}
    \bibliography{main}
}